\documentclass[fleqn]{llncs}
\usepackage[bottom]{footmisc}
\usepackage{geometry}
\usepackage{times}
\usepackage{xcolor}
\usepackage{soul}
\usepackage[utf8]{inputenc}
\usepackage[small]{caption}

\usepackage{times,amsmath,epsfig}
\usepackage{tikz}
\usepackage{floatrow}
\usepackage[ruled,vlined,linesnumbered,algo2e]{algorithm2e}
\usepackage{booktabs}
\usepackage{multirow}
\usepackage{lipsum}
\usepackage[affil-it]{authblk}
\usepackage[utf8]{inputenc}
\usepackage[english]{babel}
\usepackage{cite}
\usepackage{array}
\usepackage{amsmath}
\usepackage{footnote}
\usepackage{breqn}
\usepackage[hyphens]{url}
\usepackage{graphicx}
\usepackage{caption}
\usepackage{subcaption}
\usepackage{tabularx}
\usepackage{amsfonts}
\usepackage{calc}
\usepackage{nicefrac}       
\usepackage{microtype}      
\newcommand{\figref}[1]{Fig.~\ref{#1}}
\newfloatcommand{capbtabbox}{table}[][\FBwidth]

\title{Sequential Embedding Induced Text Clustering, a Non-parametric Bayesian Approach}

\begin{document}
	
	\author{Tiehang Duan\inst{1} \and Qi Lou\inst{2}
		 \and Sargur N. Srihari\inst{1} \and Xiaohui Xie\inst{2}}
    \institute{Department of Computer Science and Engineering\\ State University of New York at Buffalo, NY 14226, United States\\
    	\and
    	Department of Computer Science\\ University of California, Irvine, CA 92617, United States}

	\date{}
\maketitle

\begin{abstract} 
Current state-of-the-art nonparametric Bayesian text clustering methods model documents through multinomial distribution on bags of words. Although these methods can effectively utilize the word burstiness representation of documents and achieve decent performance, 
they do not explore the sequential information of text and relationships among synonyms. 
In this paper, the documents are modeled as the joint of bags of words, sequential features and word embeddings.
We proposed \underline{\textbf{S}}equential Embedding \underline{\textbf{i}}nduced \underline{\textbf{D}}irichlet \underline{\textbf{P}}rocess \underline{\textbf{M}}ixture \underline{\textbf{M}}odel (SiDPMM) to effectively exploit this joint document representation in text clustering. 
The sequential features are extracted by the encoder-decoder component. 
Word embeddings produced by the continuous-bag-of-words (CBOW) model are introduced to handle synonyms. 
Experimental results demonstrate the benefits of our model in two major aspects: 
1) improved performance across multiple diverse text datasets in terms of the normalized mutual information (NMI);
2) more accurate inference of ground truth cluster numbers with regularization effect on tiny outlier clusters.

\end{abstract}

\section{Introduction} 
The goal of text clustering is to group documents based on the content and topics. 
It has wide applications in news classification and summarization, 
document organization, trend analysis and content recommendation on social websites~\cite{Hotho01basedtext,liu2015vrca}. 
While text clustering shares the challenges of general clustering problems including high dimensionality of data, 
scalability to large datasets and prior estimation of cluster number~\cite{Berkhin2006}, 
it also bears its own uniqueness: 1) text data is inherently sequential and the order of words matters in the interpretation of document meaning. 
For example, the sentence ``people eating vegetables'' has a totally different meaning from the sentence ``vegetables eating people'', 
although two sentences share the same bag-of-words representation. 
2) Many English words have synonyms. 
Clustering methods taking synonyms into account will possibly be more effective to identify documents with similar meanings.

\begin{figure}[tbp]
	\centering
	\includegraphics[width=0.8\textwidth]{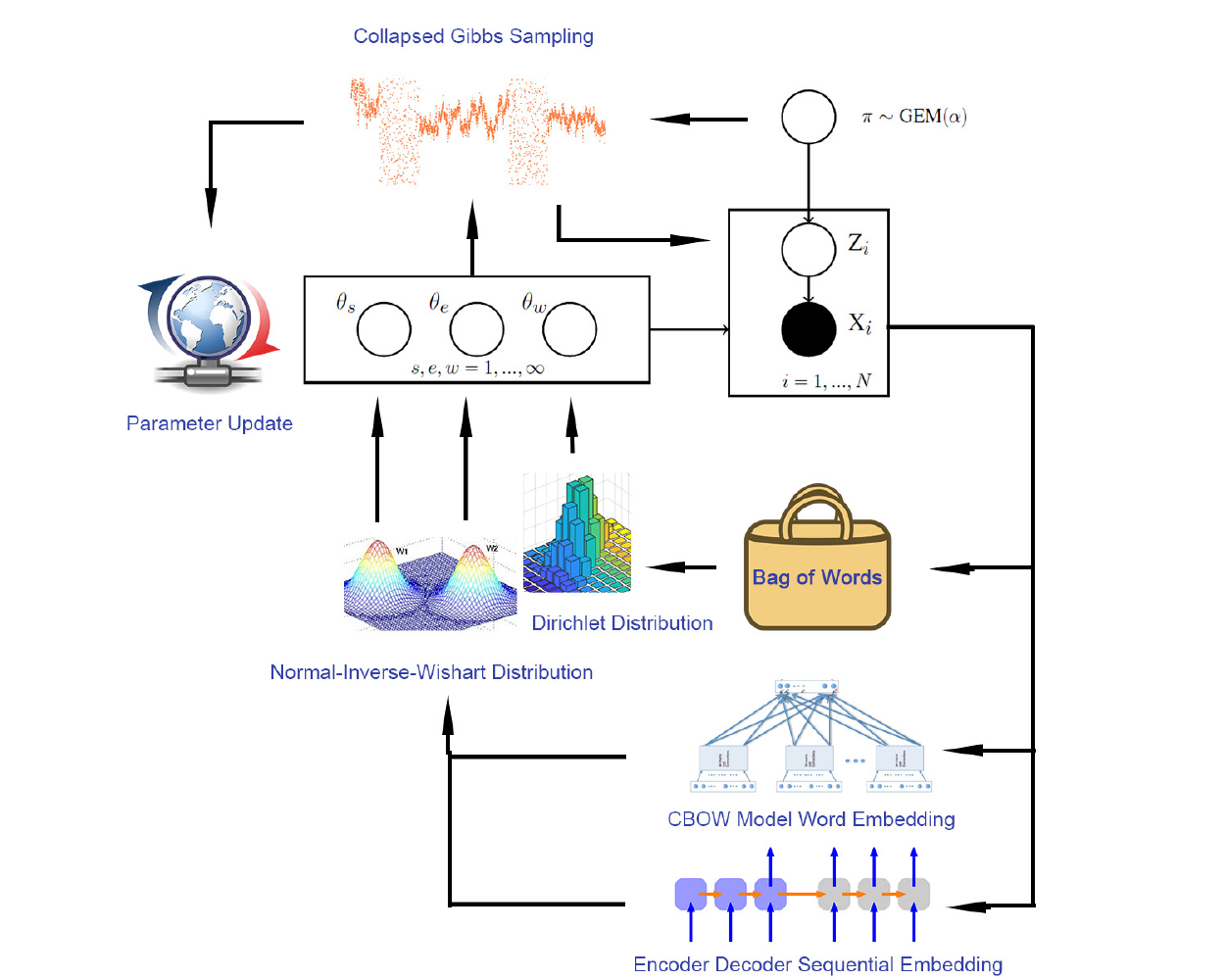}
	\caption{Illustration of the proposed sequential embedding induced Dirichlet process mixture model (SiDPMM).}
	\label{fig:sidpmm}
\end{figure}

Pioneering works in text clustering have been done to address the general challenges of clustering. 
Among them nonparametric Bayesian text clustering utilizes Dirichlet process to model the mixture distribution of text clusters 
and eliminate the need of pre-specifying the number of clusters. 
Current methods use bag of words for document modeling. 
In this work, as shown in Fig. \ref{fig:sidpmm}, the Bayesian nonparametric model is extended to utilize knowledge extracted from an encoder-decoder model and word2vec embedding, and documents are jointly modeled by bag of words, sequential features and word embeddings.
We derive an efficient collapsed Gibbs sampling algorithm for performing inference under the new model.

\subsubsection{Our Contributions.}
1) The proposed SiDPMM is able to incorporate rich feature representations.
To the best of our knowledge, this is the first work that utilizes sequential features in nonparametric Bayesian text clustering. 
The features are extracted through an encoder-decoder model. 
It also takes synonyms into account by including CBOW word embeddings as text features,
considering that documents formed with synonym words are more likely to be clustered together. 
2) We derive a collapsed Gibbs sampling algorithm for the proposed model,
which enables efficient inference.
3) Experimental results show that our model outperforms current state-of-the-art methods across multiple datasets,
and have a more accurate inference on the number of clusters due to its desirable regularization effect on tiny outlier clusters.

\section{Related Work} 
Traditional clustering algorithms such as K-means, 
Hierarchical Clustering, Singular Value Decomposition, 
Affinity Propagation have been successfully applied in the field of text clustering 
(see \cite{Rangrej2011} for a comparison of these methods on short text clustering). 
Algorithms utilizing spectral graph analysis~\cite{cai2008}, sparse matrix factorization~\cite{wang2007regu}, probabilistic models~\cite{shafiei2006,Nigam2000,Madsen2005} were proposed for performance improvement.
As text is usually represented by a huge sparse vector, 
previous works have shown that feature selection~\cite{Huang2013} and dimension reduction~\cite{GOMEZ2012} are also crucial to the task. 

Most classic methods require access to prior knowledge about the number of clusters, 
which is not always available in many real-world scenarios. 
Dirichlet Process Mixture Model (DPMM) has achieved state-of-the-art performance in text clustering 
with its capability to model arbitrary number of clusters~\cite{Yu2010,Yin2016}; 
number of clusters is automatically selected in the process of posterior inference. 
Variational inference~\cite{blei2006,ji_2017} 
and Gibbs sampling~\cite{neal2000,duan_2018} can be applied to infer cluster assignments in these models.

A closely related field of text clustering is topic modeling. 
Instead of clustering the documents, 
topic modeling aims to discover latent topics in document collections~\cite{Blei2003lda}. 
Recent works showed performance of topic modeling can be significantly improved by integrating word embeddings in the model~\cite{gao_2018_1,gao_2018_2,ijcai2017aidong,Cha2017}. 

The encoder-decoder model was recently introduced in natural language processing and computer vision to model sequential data such as phrases~\cite{gu_2017,Gu_2018, suo_2018_1, suo_2018_2} and videos~\cite{HoriHLSHM17}. It has shown great performance on a number of tasks including machine translation~\cite{cho_emnlp14}, question answering~\cite{Nie2017} and video description~\cite{HoriHLSHM17}. Its strength of extracting sequential features is revealed in these applications. 

\section{Description of SiDPMM} 
Our text clustering model is based on the Dirichlet process mixture model (DPMM),
the limit form of the Dirichlet mixture model (DMM). 
When DPMM is applied to clustering,
the size of clusters are characterized by the stick-breaking process, 
and prior of cluster assignment for each sample is characterized by the Chinese restaurant process. 
The Dirichlet process can model arbitrary number of clusters
which is typically inferred via collapsed Gibbs sampling or variational inference. 
We refer readers to~\cite{neal2000,blei2006} for more details about DPMM.

We tailor DPMM to our task by learning clusters with multiple distinct information sources for documents,
i.e., bag-of-words representations, word embeddings and sequential embeddings,
which requires specifically designed likelihood, priors, and inference mechanism.


\begin{table}[tbp]
	\centering
	\caption{Notations}
	\begin{center}
	\scalebox{0.85}{
	\begin{tabular}{cl}
		\toprule
		Notation & Meaning \\
		\midrule
		$d_{i}$ & \multicolumn{1}{p{0.45\textwidth}}{ the $i$-th document} \\
		$\vec{d}_{k,\neg i}$ & documents belonging to cluster $k$ excluding $d_{i}$ \\
		$K$ & \multicolumn{1}{p{0.45\textwidth}}{total number of clusters} \\
		$c_{i}$ & cluster assignment of $d_{i}$ \\
		$\vec{c}_{k, \neg i}$ & \multicolumn{1}{p{0.45\textwidth}}{cluster assignments of cluster $k$ excluding document i} \\
		$\theta_{k}$ & parameters of cluster $k$ \\
		$r_{k}$ & number of documents in cluster $k$ \\
		$u_{i}$ & number of words in document $i$ \\
		$u_{i}^{t}$ & occurrence of word $t$ in document $i$ \\
		$u_{k,\neg i}$ & number of words in cluster $k$ excluding $d_{i}$ \\
		$u_{k,\neg i}^{t}$ & occurrence of word $t$ in cluster $k$ excluding $d_{i}$ \\
		$w_{i}$ & the set of bag of words in $d_{i}$ \\
		$s_{i}$ & sequential information embedding of $d_{i}$ \\
		$e_{i}$ & word embedding of $d_{i}$ \\
		$V$ & vocabulary size \\
		$\Theta_{s}$ & set of hyper-parameters $\{\mu_{s}, \lambda_{s}, \nu_{s}, \Sigma_{s}\}$ \\
		$\Theta_{e}$ & set of hyper-parameters $\{\mu_{e}, \lambda_{e}, \nu_{e}, \Sigma_{e}\}$ \\
		$\alpha$ & parameter of Chinese restaurant process\\
		$\beta$ & hyper-parameter for multinomial modeling of bag of words \\
		$\epsilon$ & dimensionality of sequential embedding vector \\
		$\delta_k$ & parameter of multinomial distribution for the $k$-th cluster \\
		\bottomrule
	\end{tabular}%
    }
    \end{center}
	\label{tab:addlabel}%
\end{table}
To start with, we first introduce the likelihood function $\boldmath{F}(d_{i}|\theta_{k})$ over documents:
\begin{align} \label{eq:4}
\boldmath{F} (d_{i}|\theta_{k})=
\text{Mult}(w_{i}|\delta_{k}) \mathcal{N}(e_{i}|\mu_{e}^{k},\Sigma_{e}^{k})
\mathcal{N}(s_{i}|\mu_{s}^{k},\Sigma_{s}^{k})
\end{align}
where $\theta_{k}=(\mu_{e}^k, \Sigma_{e}^k, \mu_{s}^k, \Sigma_{s}^k, \delta_k)$, 
with $\delta_{k} = (\delta_{k}^{1}, \ldots, \delta_{k}^V)$ and $\sum_{j=1}^{V}\delta_{k}^{j}=1$. 
$e_{i}$ is the word embedding and $s_{i}$ is the encoded sequential vector. 
The multinomial component $\text{Mult}(w_{i}|\delta_{k})$ captures the distribution of bag of words; 
the Normal components $\mathcal{N}(e_{i}|\mu_{e}^{k},\Sigma_{e}^{k})$,
$\mathcal{N}(s_{i}|\mu_{s}^{k},\Sigma_{s}^{k})$ measure similarities of word and sequential embeddings. 
This model is general enough to model the characteristic of any text 
and also specific enough to capture the key information of each document including word embeddings and sequential embeddings.

 The prior is set to be conjugate with the likelihood for integrating out the cluster parameters during the inference phase. As Dirichlet distribution is the conjugate prior of multinomial distribution and Normal-inverse-Wishart(NiW) is the conjugate prior of normal distribution, we used the composition of Dirichlet distribution and NiW distribution to serve as the conjugate prior $\mathbb{G}_{0}$, which is defined as:

\begin{align} \label{eq:5}
\mathbb{G}_{0}(\theta_{k})=\text{Diri}(\delta_{k}|\beta)\text{NiW}(\mu_{s}^{k}, \Sigma_{s}^{k}|\Theta_{s})\text{NiW}(\mu_{e}^{k}, \Sigma_{e}^{k}|\Theta_{e})
\end{align}

where $\text{Diri}$ denotes the Dirichlet distribution and $\text{NiW}$ denotes the Normal-inverse-Wishart distribution.  $\Theta_{s}$ denotes hyper-parameters $\{\mu_{s0},\lambda_{s0}, \nu_{s0}, \Sigma_{s0}\}$ for the encoder-decoder component and $\Theta_{e}$ denotes hyper-parameters $\{\mu_{e0},\lambda_{e0}, \nu_{e0}, \Sigma_{e0}\}$ for CBOW word embedding component.

\section{Inference via Collapsed Gibbs Sampling} 
We adopt collapsed Gibbs sampling for inference due to its efficiency. 
It reduces the dimensionality of the sampling space by integrating out cluster parameters, which leads to faster convergence.  

The cluster assignment $k$ for document $i$ is decided based on the posterior distribution $p(c_{i}=k|\vec{c}_{\neg i}, \vec{d},\theta)$. It can be represented as product of cluster prior and document likelihood.
\begin{align} \label{eq:9}
\begin{split}
&p(c_{i}|\vec{c}_{\neg i}, \vec{d},\theta)=\frac{p(c_{i},\vec{c}_{\neg i},\vec{d}|\theta)}{p(\vec{c}_{\neg i},\vec{d}|\theta)} \propto \frac{p(\vec{c},\vec{d}|\theta)}{p(\vec{c}_{\neg i},\vec{d}_{\neg i}|\theta)}
=\frac{p(\vec{c}|\theta)}{p(\vec{c}_{\neg i}|\theta)}\frac{p(\vec{d}|\vec{c},\theta)}{p(\vec{d}_{\neg i}|\vec{c},\theta)}
\\&=p(c_{i}|\vec{c}_{\neg i},\theta)p(d_{i}|\vec{d}_{\neg i},\vec{c},\theta)
\end{split}
\end{align}
Based on the Chinese restaurant process depiction of DPMM, 
we have 
\begin{equation} \label{eq:11}
\begin{split}
p(c_{i}|\vec{c}_{\neg i}, \theta)
&=p(c_{i}|\vec{c}_{\neg i}, \alpha)\\
&=
\begin{cases} 
\frac{r_{k,\neg i}}{D-1+\alpha}\,\,\text{ choose an existing cluster $k$}\\
\frac{\alpha}{D-1+\alpha}\,\,\text{ create a new cluster}
\end{cases}
\end{split}
\end{equation}
$(D-1)$ is the total number of documents in the corpus excluding current document $i$.

Given the number of variables introduced in the model, 
direct sampling from the joint distribution is not practical.
Thus, we assume conditional independence on the variables by allowing the factorization of the second term in \eqref{eq:9} as: 
\begin{equation} \label{eq:12}
\begin{split}
p(d_{i}|\vec{d}_{\neg i},\vec{c},\theta)
\propto p(w_{i}|\vec{d}_{\neg i},\vec{c},\theta) p(e_{i}|\vec{d}_{\neg i},\vec{c},\theta) p(s_{i}|\vec{d}_{\neg i},\vec{c},\theta)
\end{split}
\end{equation}
The calculation for each component $p(w_{i}|\vec{d}_{\neg i},\vec{c},\theta)$, 
$p(e_{i}|\vec{d}_{\neg i},\vec{c},\theta)$ 
and $p(s_{i}|\vec{d}_{\neg i},\vec{c},\theta)$ is derived below:
\begin{equation} \label{eq:13}
\begin{split}
p(w_{i}|\vec{d}_{\neg i},\vec{c},\theta)
=p(w_{i}|c_{i} = k,\vec{d}_{k,\neg i},\beta)
=\int p(w_{i}|\delta_{k})p(\delta_{k}|\vec{d}_{k,\neg i},\beta)d\delta_{k} 
\end{split}
\end{equation}
where the first term in the above integral is
\begin{equation} \label{eq:14}
p(w_{i}|\delta_{k})=\prod_{t\in w_{i}}\text{Mult}(t|\delta_{k})=\prod_{t=1}^{V}\delta_{k,t}^{u_{i}^{t}}
\end{equation}
$\delta_{k,t}$ is the probability of term $t$ bursting in cluster $k$ and $u_{i}^{t}$ is the count of term $t$ in document $i$.
The second term in (\ref{eq:13}) is
\begin{equation} \label{eq:16}
\begin{split}
p(\delta_{k}|\vec{d}_{k, \neg i},\beta)=\frac{p(\delta_{k}|\beta) p(\vec{d}_{k,\neg i}|\delta_{k})}{\int_{k} p(\delta_{k}|\beta) p(\vec{d}_{k,\neg i}|\delta_{k})d\delta_{k}}
\end{split}
\end{equation}
By defining $\Delta(\vec{\beta})=\frac{\prod_{k=1}^{K}\Gamma(\beta)}{\Gamma(\sum_{k=1}^{K}\beta)}$ similar to \cite{Yin2016}, 
we have
\begin{equation} \label{eq:17}
\begin{split}
&p(\delta_{k}|\vec{d}_{k,\neg i},\beta)=\frac{\frac{1}{\Delta(\beta)}\prod_{t=1}^{V}\delta_{k,t}^{\beta-1}\prod_{t=1}^{V}\delta_{k,t}^{u_{k,\neg i}^{t}}}{\int_{k} \frac{1}{\Delta(\beta)}\prod_{t=1}^{V}\delta_{k,t}^{\beta-1}\prod_{t=1}^{V}\delta_{k,t}^{u_{k,\neg i}^{t}}d\delta_{k}}\\ &=\frac{1}{\Delta(\vec{u}_{k,\neg i}+\beta)}\prod_{t=1}^{V}\delta_{k,t}^{u_{k,\neg i}^{t}+\beta-1}
\end{split}
\end{equation}
Based on (\ref{eq:14}) and (\ref{eq:17}), (\ref{eq:13}) becomes
\begin{equation} \label{eq:18}
	\begin{split}
	&p(w_{i}|\vec{d}_{\neg i},\vec{c},\theta)=\int_{k} \frac{1}{\Delta(\vec{u}_{k,\neg i}+\beta)}\prod_{t=1}^{V}\delta_{k,t}^{u_{k,\neg i}^{t}+\beta-1}\prod_{t=1}^{V}\delta_{k,t}^{u_{i}^{t}}d\delta_{k}
	\\&=\frac{\Delta(\vec{u}_{k}+\beta)}{\Delta(\vec{u}_{k,\neg i}+\beta)}=\frac{\prod_{t=1}^{V}\Gamma(u_{k}^{t}+\beta)}{\Gamma(\sum_{t=1}^{V}(u_{k}^{t}+\beta))}\frac{\Gamma(\sum_{t=1}^{V}(u_{k,\neg i}^{t}+\beta))}{\prod_{t=1}^{V}\Gamma(u_{k, \neg i}^{t}+\beta)}\\
	&=\frac{\prod_{t=1}^{V}\prod_{j=1}^{u_{i}^{t}}(u_{k,\neg i}^{t}+\beta+j-1)}{\prod_{j=1}^{u_{i}}(u_{k,\neg i}+V\beta+j-1)}
	\end{split}
\end{equation}
As we see from \eqref{eq:18}, the high dimensionality challenge of text clustering is naturally circumvented 
by multiplying one dimension of the vector space at a time.
$p(e_{i}|\vec{d}_{\neg i},\vec{c},\theta)$ and $p(s_{i}|\vec{d}_{\neg i},\vec{c},\theta)$ 
in \eqref{eq:12} are derived based on properties of NiW distribution:
\begin{equation} \label{eq:19}
\begin{split}
&p(s_{i}|\vec{d}_{\neg i},\vec{c},\theta) =p(s_{i}|c_{i}=k,\vec{d}_{k,\neg i},\theta)\\ 
&=\int_{\mu_{k}} \int_{\Sigma_{k}}p(s_{i}|\mu_{k}, \Sigma_{k})p(\mu_{k},\Sigma_{k}|c_{i}=k,\vec{d}_{k,\neg i},\theta)d\mu_{k} d\Sigma_{k}\\
&=\int_{\mu_{k}} \int_{\Sigma_{k}} \mathcal{N} (s_{i}|\mu_{k}, \Sigma_{k}) \text{NiW}(\mu_{k}, \Sigma_{k}|\Theta_{s}^{k,\neg i})d\mu_{k} d\Sigma_{k}
\end{split}
\end{equation}
where $\mu$ and $\Sigma$ are the mean and variance of the sequential embedding, $\Theta_{s}^{k,\neg i}$ includes $\{ \mu_{s}^{k,\neg i},\lambda_{s}^{k,\neg i},\nu_{s}^{k,\neg i},\Sigma_{s}^{k,\neg i}\}$ which is the hyper-parameter in the NiW distribution of cluster $k$.

We define the normalization constant $Z(\epsilon, \lambda, \nu, \Sigma)$ of NiW distribution as 
\begin{equation} \label{eq:26}
Z(\epsilon, \lambda, \nu, \Sigma)=2^{\frac{(\nu+1)\epsilon}{2}}\pi^{\frac{\epsilon(\epsilon+1)}{4}}\lambda^{\frac{-\epsilon}{2}}|\Sigma|^{\frac{-\nu}{2}}\prod_{i=1}^{\epsilon}\Gamma(\frac{\nu+1-i}{2})
\end{equation}
where $\epsilon$ is the dimensionality of sequential embedding vector. Therefore
\begin{equation} \label{eq:22}
\begin{split}
&p(s_{i}|\vec{d}_{\neg i},\vec{c},\theta)\\&=\int_{\mu_{k}} \int_{\Sigma_{k}} \mathcal{N} (s_{i}|\mu_{k}, \Sigma_{k}) \text{NiW}(\mu_{k}, \Sigma_{k}|\Theta_{s}^{k,\neg i})d\mu_{k} d\Sigma_{k}\\& =(2\pi)^{\frac{-\epsilon}{2}}\frac{Z(\epsilon,\lambda_{s}^{k},\nu_{s}^{k},\Sigma_{s}^{k})}{Z(\epsilon,\lambda_{s}^{k, \neg i},\nu_{s}^{k, \neg i},\Sigma_{s}^{k, \neg i})}\\& =(\pi)^{\frac{-\epsilon}{2}}(\frac{\lambda_{s}^{k}}{\lambda_{s}^{k,\neg i}})^{\frac{-\epsilon}{2}}\frac{|\Sigma_{s}^{k}|^{\frac{-\nu_{s}^{k}}{2}}}{|\Sigma_{s}^{k,\neg i}|^{\frac{-\nu_{s}^{k,\neg i}}{2}}}\prod_{j=1}^{\epsilon}\frac{\Gamma(\frac{\nu_{s}^{k}+1-j}{2})}{\Gamma(\frac{\nu_{s}^{k,\neg i}+1-j}{2})}
\end{split}
\end{equation}
As
$\nu_{s}^{k}=\nu_{s}^{k,\neg i}+1$,
we have
\begin{equation} \label{eq:23}
\begin{split}
p(s_{i}|\vec{d}_{\neg i},\vec{c},\theta) =(\pi)^{\frac{-\epsilon}{2}}(\frac{\lambda_{s}^{k}}{\lambda_{s}^{k,\neg i}})^{\frac{-\epsilon}{2}}\frac{|\Sigma_{s}^{k}|^{\frac{-\nu_{s}^{k}}{2}}}{|\Sigma_{s}^{k,\neg i}|^{\frac{-\nu_{s}^{k,\neg i}}{2}}}\frac{\Gamma(\frac{\nu_{s}^{k}}{2})}{\Gamma(\frac{\nu_{s}^{k}-\epsilon}{2})}
\end{split}
\end{equation}
The derivation of $p(e_{i}|\vec{d}_{\neg i},\vec{c},\theta)$ is analogous to that of $p(s_{i}|\vec{d}_{\neg i},\vec{c},\theta)$ 
as they are following the same form of distribution, thus,
\begin{equation} \label{eq:24}
\begin{split}
p(e_{i}|\vec{d}_{\neg i},\vec{c},\theta) =(\pi)^{\frac{-\epsilon}{2}}(\frac{\lambda_{e}^{k}}{\lambda_{e}^{k,\neg i}})^{\frac{-\epsilon}{2}}\frac{|\Sigma_{e}^{k}|^{\frac{-\nu_{e}^{k}}{2}}}{|\Sigma_{e}^{k,\neg i}|^{\frac{-\nu_{e}^{k,\neg i}}{2}}}\frac{\Gamma(\frac{\nu_{e}^{k}}{2})}{\Gamma(\frac{\nu_{e}^{k}-\epsilon}{2})}
\end{split}
\end{equation}

Algorithm~\ref{algo_cgs} presents the complete inference procedure.

\begin{algorithm2e}[!ht]
\caption{Inference of SiDPMM Model}
\label{algo_cgs}
\SetAlgoLined
\SetKwInOut{Input}{Data}
\SetKwInOut{Output}{Result}
\Input{For each document $i$, the bag of words $\vec{w_{i}}$, word embedding $\vec{e_{i}}$, sequential embedding~$\vec{s_{i}}$} 

\Output{Number of clusters $K$, cluster assignments for each document $\vec{c}$} 

\tcc{Initialization}
K=0

\For{each document $i$}{
	
	compute cluster prior $p(c_{i}|\vec{c}_{\neg i}, \alpha)$ $\vartriangleright$~\eqref{eq:11}
	
	calculate $p(w_{i}|\vec{d}_{k, \neg i},c_{i}=k,\theta)$ $\vartriangleright$~\eqref{eq:18}
	
	calculate $p(s_{i}|\vec{d}_{k, \neg i},c_{i}=k,\theta)$ $\vartriangleright$~\eqref{eq:23}
	
	calculate $p(e_{i}|\vec{d}_{k, \neg i},c_{i}=k,\theta)$ $\vartriangleright$~\eqref{eq:24}
		
	calculate $p(d_{i}|\vec{d}_{k,\neg i},c_{i}=k,\theta)$ $\vartriangleright$~\eqref{eq:12}
	
	sample cluster $c_{i} \!\sim\! p(c_{i}=k|\vec{c}_{\neg i}, \vec{d},\theta)$ $\vartriangleright$~\eqref{eq:9}
	
	\If{$c_{i}=K+1$}{
		K=K+1
		}
	
	update parameters of cluster $c_{i}$
	
	}

\tcc{Collapsed Gibbs Sampling, N iterations}

\For{Iter= 1 to N}{
	\For{each document $i$}{
		delete document $i$ from cluster $c_{i}$, update parameters of cluster $c_{i}$ 
		
		\If{cluster $c_{i}$ is empty}{K=K-1}
		
		repeat line 3 to line 7
		
		sample a new cluster $c_{i}$ for document $i$ $\vartriangleright\,$ (\ref{eq:9})
		
		\If{$c_{i}=K+1$}{
			K=K+1
		}
		
		update parameters of cluster $c_{i}$
		
		}
	}
\end{algorithm2e}

\section{Extraction of Sequential Feature and Synonyms Embedding} 
In this section, we describe how to extract sequential embeddings with an encoder-decoder component and synonyms embeddings with the CBOW model.

\begin{figure}[tbp]
	\centering
	\begin{subfigure}[b]{0.6\textwidth}
	\centering
	
	\begin{tikzpicture}[scale=.32, transform shape]
	\tikzstyle{every node} = [scale=1,shape=rectangle, rounded corners,draw, align=center,fill=gray!10]
	\draw (-12, 3) rectangle (-4, 4);
	\node (nn) at (-8, 3.5)[opacity=0,text opacity=1] {\LARGE LSTM Layer};
	\draw (-12, -1) rectangle (-4, 0);
	\node (nn) at (-8, -0.5)[opacity=0,text opacity=1] {\LARGE LSTM Layer};
	\draw (-12, 1) rectangle (-4, 2);
	\node (nn) at (-8, 1.5)[opacity=0,text opacity=1] {\LARGE LSTM Layer};
	\draw (0, 3) rectangle (8, 4);
	\node (nn) at (4, 3.5)[opacity=0,text opacity=1] {\LARGE LSTM Layer};
	\draw (0, -1) rectangle (8, 0);
	\node (nn) at (4, -0.5)[opacity=0,text opacity=1] {\LARGE LSTM Layer};
	\draw (0, 1) rectangle (8, 2);
	\node (nn) at (4, 1.5)[opacity=0,text opacity=1] {\LARGE LSTM Layer};
	\draw[thick] (-11,4) -- (-11,5.5);
	\draw[thick] (-9,4) -- (-9,5.5);
	\draw[thick] (-7,4) -- (-7,5.5);
	\draw[thick] (-5,4) -- (-5,5.5);
	
	\draw[thick] (1,4) -- (1,5.5);
	\draw[thick] (3,4) -- (3,5.5);
	\draw[thick] (5,4) -- (5,5.5);
	\draw[thick] (7,4) -- (7,5.5);
	
	\draw[thick] (-11,-2.5) -- (-11,-1);
	\draw[thick] (-9,-2.5) -- (-9,-1);
	\draw[thick] (-7,-2.5) -- (-7,-1);
	\draw[thick] (-5,-2.5) -- (-5,-1);
	
	\draw[thick] (1,-2.5) -- (1,-1);
	\draw[thick] (3,-2.5) -- (3,-1);
	\draw[thick] (5,-2.5) -- (5,-1);
	\draw[thick] (7,-2.5) -- (7,-1);
	
	\draw[thick] (-11,0) -- (-11,1);
	\draw[thick] (-9,0) -- (-9,1);
	\draw[thick] (-7,0) -- (-7,1);
	\draw[thick] (-5,0) -- (-5,1);
	
	\draw[thick] (-11,2) -- (-11,3);
	\draw[thick] (-9,2) -- (-9,3);
	\draw[thick] (-7,2) -- (-7,3);
	\draw[thick] (-5,2) -- (-5,3);
	
	\draw[thick] (1,0) -- (1,1);
	\draw[thick] (3,0) -- (3,1);
	\draw[thick] (5,0) -- (5,1);
	\draw[thick] (7,0) -- (7,1);
	
	\draw[thick] (1,2) -- (1,3);
	\draw[thick] (3,2) -- (3,3);
	\draw[thick] (5,2) -- (5,3);
	\draw[thick] (7,2) -- (7,3);
	
	\draw[line width=1pt,->] (-5, 5.5) .. controls(-2.5,5.0) and (-2.5,4.0) .. (0, 3.5);
	
	\draw[line width=1pt,->] (-5, 3) .. controls(-2.5,2.5) and (-2.5,2.0) .. (0, 1.5);
	
	\draw[line width=1pt,->] (-5, 1) .. controls(-2.5,0.5) and (-2.5,0) .. (0, -0.5);
	
	\draw (-6, 6.5) -- (-4, 6.5) node[minimum size=1cm,pos=.5,sloped,below] {\LARGE Encoded Sequential Vector};
	
	\node[minimum size=1cm] (X1) at (-11, -2.5) {\LARGE A};
	\node[minimum size=1cm] (X2) at (-9, -2.5) {\LARGE B};
	\node[minimum size=1cm] (X3) at (-7, -2.5) {\LARGE C};
	\node[minimum size=1cm] (X4) at (-5, -2.5) {\LARGE D};
	
	\node[minimum size=1cm] (Y1) at (1, -2.5) {\LARGE $<eos>$};
	\node[minimum size=1cm] (Y2) at (3, -2.5) {\LARGE X};
	\node[minimum size=1cm] (Y3) at (5, -2.5) {\LARGE Y};
	\node[minimum size=1cm] (Y4) at (7, -2.5) {\LARGE Z}; 	
	
	\node[minimum size=1cm] (Z1) at (1, 5.5) {\LARGE X};
	\node[minimum size=1cm] (Z2) at (3, 5.5) {\LARGE Y};
	\node[minimum size=1cm] (Z3) at (5, 5.5) {\LARGE Z};
	\node[minimum size=1cm] (Z4) at (7, 5.5) {\LARGE $<eos>$}; 
	
	\draw (-11, 5.5) -- (-7, 5.5) node[minimum size=1cm,pos=.5,sloped,below] {\LARGE Encoder Hidden State Vectors};	
	
	\end{tikzpicture}
	\caption{}
	\label{Figure 12}

	\end{subfigure}%
	\begin{subfigure}[b]{0.4\textwidth}
	\centering
	\includegraphics[width=0.85\textwidth]{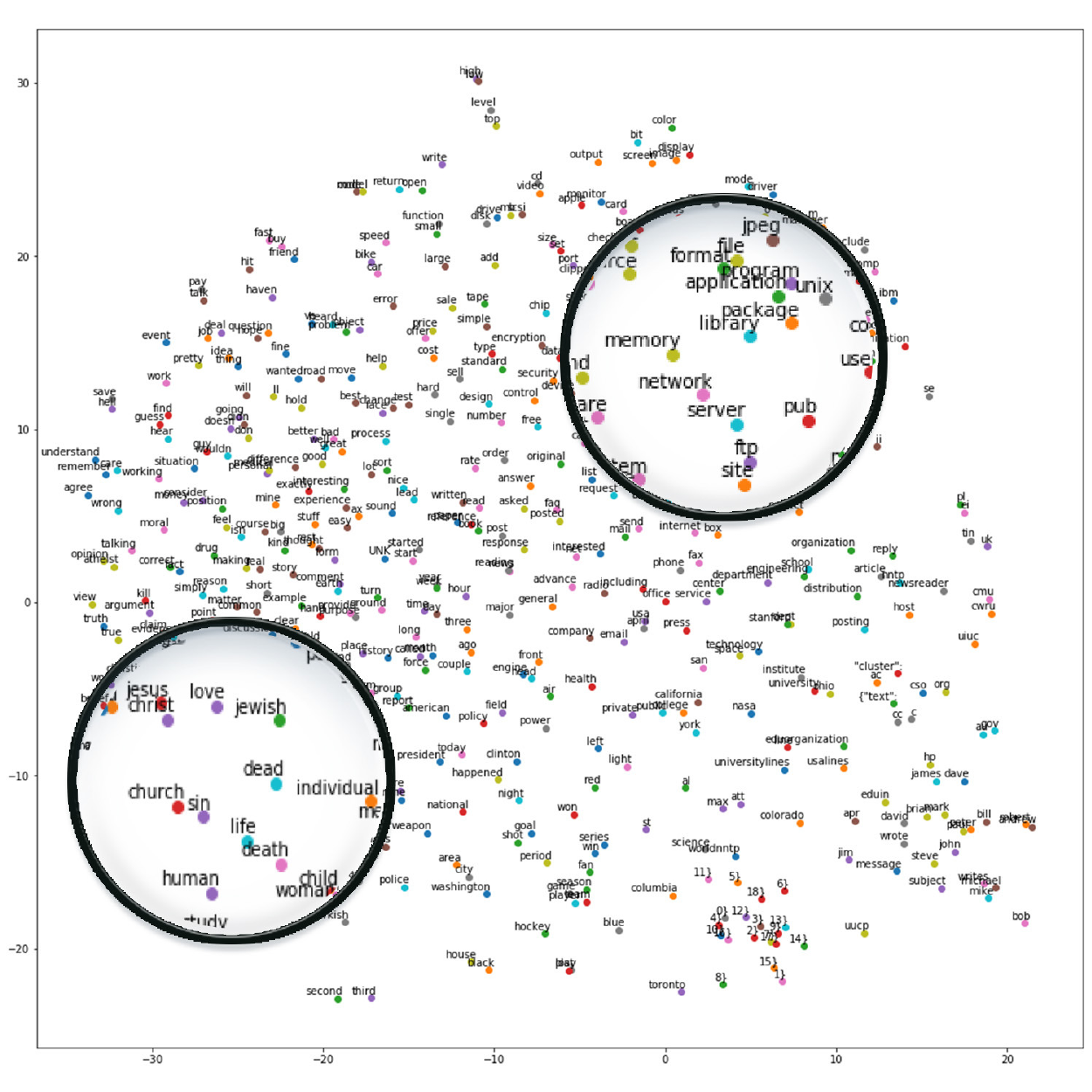}
	\caption{}
	\label{Figure 2}
	\end{subfigure}%
	\caption{(a) The Encoder-Decoder Component. 
		It is formed by two LSTM stacks, one is for mapping a sequential input data to a fixed-length vector, 
		the other is for decoding the vector to a sequential output. (b) Word embedding of Google News Title Set. Words describing the same topic have similar embeddings and are clustered together}
	\label{fig:22}
\end{figure}
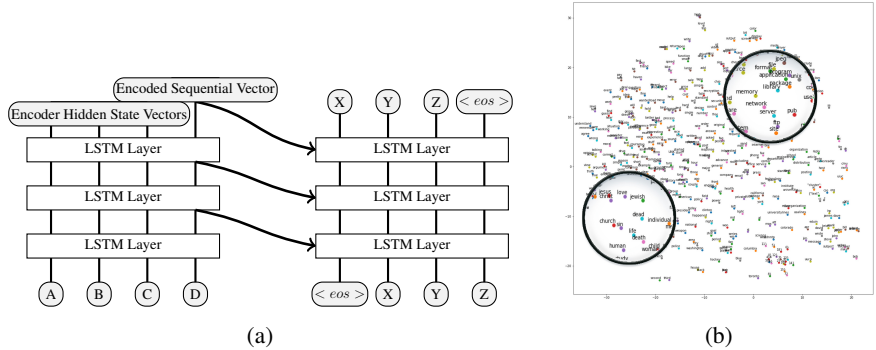

The encoder-decoder component is formed with two LSTM stacks \cite{duan_2017}, 
one is for mapping the sequential input data to a fixed length vector, 
the other is for decoding the vector to a sequential output. 
To learn embeddings, 
we set the input sequence and output sequence to be the same. 
An illustration of the encoder-decoder mechanism is shown in \figref{Figure 12}. 
The last output of the encoder LSTM stack contains information of the whole phrase. 
In machine translation, researchers have found the information is rich enough for the original phrase to be decoded into translations of another language~\cite{LuongPM15}.

 Current state-of-the-art text clustering methods adopt one-hot encoding for word representation. 
 It neglects semantic relationship between similar words. 
 Recently, researchers have shown multiple degrees of similarity can be revealed among words with word embedding techniques~\cite{mikolov2013linguistic}. 
 Utilizing such embeddings means we can cluster the documents based on \emph{meaning of words} instead of the word itself. 
 As shown in Fig. \ref{Figure 2}, words describing the same topic have similar embeddings and are clustered together. 
 The CBOW model is used to learn word embeddings by predicting each word based on word context (weighted nearby surrounding words). 
 The embedding vector $e_{i}$ is the average of word embeddings in $d_{i}$. 
 Readers are referred to \cite{Mikolov2013} for details about the CBOW model.

\section{Experiments} 
In this section, we will demonstrate the effectiveness of our approach through a series of experiments.
The detailed experimental settings are as follows:

\subsubsection{Datasets}
 We run experiments on four diverse datasets including 20 News Group (20NG)\footnote{\url{http://qwone.com/~jason/20Newsgroups/}}, 
 Tweet Set\footnote{\url{http://trec.nist.gov/data/microblog.html}}, 
 and two datasets from~\cite{Yin2016}: Google News Title Set (T-Set) and Google News Snippet Set~(S-Set). 
 The 20NG dataset contains long documents with an average length of 138 while the documents in T-Set and Tweet Set are short with average length less than 10. 
 Phrase structures are sparse in T-Set, while rich in 20NG and S-Set. 
 The Tweet Set contains moderate phrase structures.
 

\subsubsection{Baselines}
We compare SiDPMM against two classic clustering methods, K-means and latent Dirichlet allocation (LDA),
and two recent methods GSDMM and GSDPMM that are state-of-the-art in nonparametric Bayesian text clustering.

\subsubsection{Metrics}
We take the normalized mutual information (NMI) 
as the major evaluation metric in our experiments
since NMI is widely used in this field.
NMI scores range from 0 to 1. 
Perfect labeling is scored to 1 while random assignments tend to achieve scores close to 0. 

\begin{table}[htbp]
	\centering
	\caption{NMI scores on various dataset-parameter settings. 
		$K$ is the prior number of clusters for K-means, LDA and GSDMM, 
		set to be four different values including the ground truth for each dataset. 
		$K$ is not used for SiDPMM and GSDPMM. 
		20 independent runs for each setting.
	}
	\scalebox{0.85}{
		\begin{minipage}{\textwidth}
			\begin{tabular}{lcccccccc}
				\toprule
				& K     & SiDPMM & SiDPMM-sf\footnote{SiDPMM model only integrating sequential features} & SiDPMM-we\footnote{SiDPMM model only integrating word embeddings} & K-means & LDA   & GSDMM & GSDPMM \\
				\midrule
				\multirow{4}[2]{*}{20NG} & 10    & .689$\pm$.006 & .686$\pm$.005 & .680$\pm$.006 & .235$\pm$.008 & .585 $\pm$ .013 & .613 $\pm$ .007 & .667 $\pm$ .004 \\
				& 20    & .689$\pm$.006 & .686$\pm$.005 & .680$\pm$.006 & .321$\pm$.006 & .602 $\pm$ .012 & .642 $\pm$ .004 & .667 $\pm$ .004 \\
				& 30    & .689$\pm$.006 & .686$\pm$.005 & .680$\pm$.006 & .336$\pm$.005 & .611 $\pm$ .012 & .649 $\pm$ .005 & .667 $\pm$ .004 \\
				& 50    & .689$\pm$.006 & .686$\pm$.005 & .680$\pm$.006 & .348$\pm$.006 & .617 $\pm$ .013 & .656 $\pm$ .002 & .667 $\pm$ .004 \\
				\midrule
				\multirow{4}[2]{*}{T-Set} & 100   & .878$\pm$.003 & .872$\pm$.003 & .877$\pm$.005 & .687$\pm$.005 & .769 $\pm$ .012 & .830 $\pm$ .004 & .873 $\pm$ .002 \\
				& 150   & .878$\pm$.003 & .872$\pm$.003 & .877$\pm$.005 & .721$\pm$.009 & .784 $\pm$ .015 & .852 $\pm$ .009 & .873 $\pm$ .002 \\
				& 152   & .878$\pm$.003 & .872$\pm$.003 & .877$\pm$.005 & .720$\pm$.007 & .786 $\pm$ .014 & .853 $\pm$ .009 & .873 $\pm$ .002 \\
				& 200   & .878$\pm$.003 & .872$\pm$.003 & .877$\pm$.005 & .730$\pm$.008 & .806 $\pm$ .013 & .868 $\pm$ .006 & .873 $\pm$ .002 \\
				\midrule
				\multirow{4}[2]{*}{S-Set} & 100   & .916$\pm$.004 & .910$\pm$.005 & .902$\pm$.003 & .739$\pm$.006 & .848 $\pm$ .005 & .854 $\pm$ .004 & .891 $\pm$ .004 \\
				& 150   & .916$\pm$.004 & .910$\pm$.005 & .902$\pm$.003 & .756$\pm$.006 & .850 $\pm$ .006 & .867 $\pm$ .008 & .891 $\pm$ .004 \\
				& 152   & .916$\pm$.004 & .910$\pm$.005 & .902$\pm$.003 & .757$\pm$.007 & .852 $\pm$ .005 & .867 $\pm$ .009 & .891 $\pm$ .004 \\
				& 200   & .916$\pm$.004 & .910$\pm$.005 & .902$\pm$.003 & .768$\pm$.007 & .862 $\pm$ .004 & .885 $\pm$ .005 & .891 $\pm$ .004 \\
				\midrule
				\multirow{4}[2]{*}{Tweet} & 50    & .894$\pm$.007 & .887$\pm$.006 & .884$\pm$.005 & .696$\pm$.008 & .775 $\pm$ .012 & .844 $\pm$ .006 & .875 $\pm$ .005 \\
				& 90    & .894$\pm$.007 & .887$\pm$.006 & .884$\pm$.005 & .725$\pm$.007 & .797 $\pm$ .011 & .862 $\pm$ .008 & .875 $\pm$ .005 \\
				& 110   & .894$\pm$.007 & .887$\pm$.006 & .884$\pm$.005 & .732$\pm$.006 & .806 $\pm$ .010 & .867 $\pm$ .006 & .875 $\pm$ .005 \\
				& 150   & .894$\pm$.007 & .887$\pm$.006 & .884$\pm$.005 & .742$\pm$.006 & .811 $\pm$ .012 & .871 $\pm$ .004 & .875 $\pm$ .005 \\
				\bottomrule
			\end{tabular}%
		\end{minipage}
	}
	\label{Table 1}%
\end{table}%

\subsubsection{Encoder-decoder component} We truncate the sequence length to be 48 for Tweet Set and Google News dataset and 240 for 20NG dataset. The document with characters length shorter than this sequence length is padded with zeros. The encoder-decoder model is trained for 10 iterations. 
The length of hidden vectors is set to be 40, 
and length of input vector is 67 (number of different characters). 
Weights in the LSTM stack are uniformly initialized to be 0.01. 
Adam~\cite{KingmaB14} optimizer is used to optimize the network with its learning rate set to 0.01. 

\subsubsection{Word embedding component} The vocabulary size is set to 100,000 which is enough to accommodate most of the words present in the dataset.
We set the embedding vector length to be 40. 
To facilitate training with small datasets such as the Tweet Set, 
we augment each dataset with a well-known large-scaled text dataset\footnote{\url{http://mattmahoney.net/dc/text8.zip}} during training. 
Window size is set to be 1, meaning we only consider the words that are neighbors of the target word as its word context. 
We apply stochastic gradient descent for optimization with a total of 100,000 descent steps. 

\subsubsection{Priors} Hyper-parameter $\alpha$ of the Dirichlet process is set to be $0.1\times|\vec{d}|$, where $|\vec{d}|$ is number of documents in the dataset.
Hyper-parameter $\beta$ for the Multinomial modeling of bag of words is $0.002\times V$, 
and parameters for the prior NiW distribution of word embedding and sequential embedding are 
$\{\mu_{0}=\vec{0},\lambda_{0}=1,\nu_{0}=\epsilon,\Sigma_{0}=I\}$.

\subsection{Empirical Results}
Table~\ref{Table 1} reports the mean and standard deviation of the NMI scores across various settings. 
From Table~\ref{Table 1}, 
we observe that SiDPMM outperforms K-means, LDA and GSDMM across all the settings by significant margins. 
GSDPMM has comparative performance with SiDPMM on T-Set, while SiDPMM performs better in other three datasets. We noted the average length of T-Set is short and phrase structures are scarce in its documents. To unveil the influence of each of the component on the model performance, we included implementation of SiDPMM model only integrating sequential features (denoted as SiDPMM-sf) and SiDPMM model only integrating word embeddings (denoted as SiDPMM-we) into the comparison. We noted the contribution from sequential embedding is significant in 20NG, S-Set and moderate in Tweet-Set.

\begin{figure}[tbp]
	\centering
	\begin{subfigure}[b]{0.4\textwidth}
		\centering
		\includegraphics[width=\textwidth]{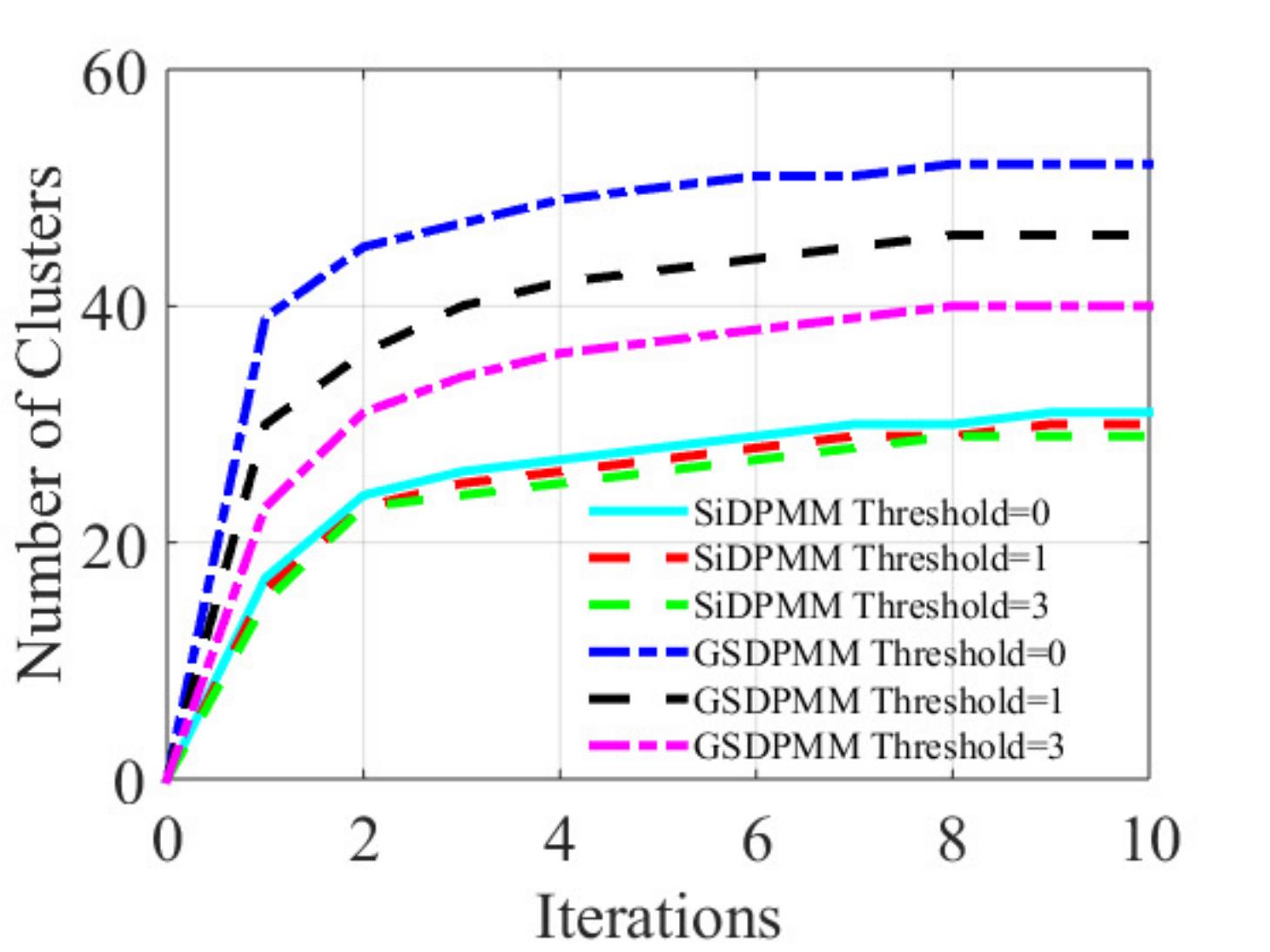}
		\caption{20NG}
		\label{Fig8_1}
	\end{subfigure}%
	\begin{subfigure}[b]{0.4\textwidth}
		\centering
		\includegraphics[width=\textwidth]{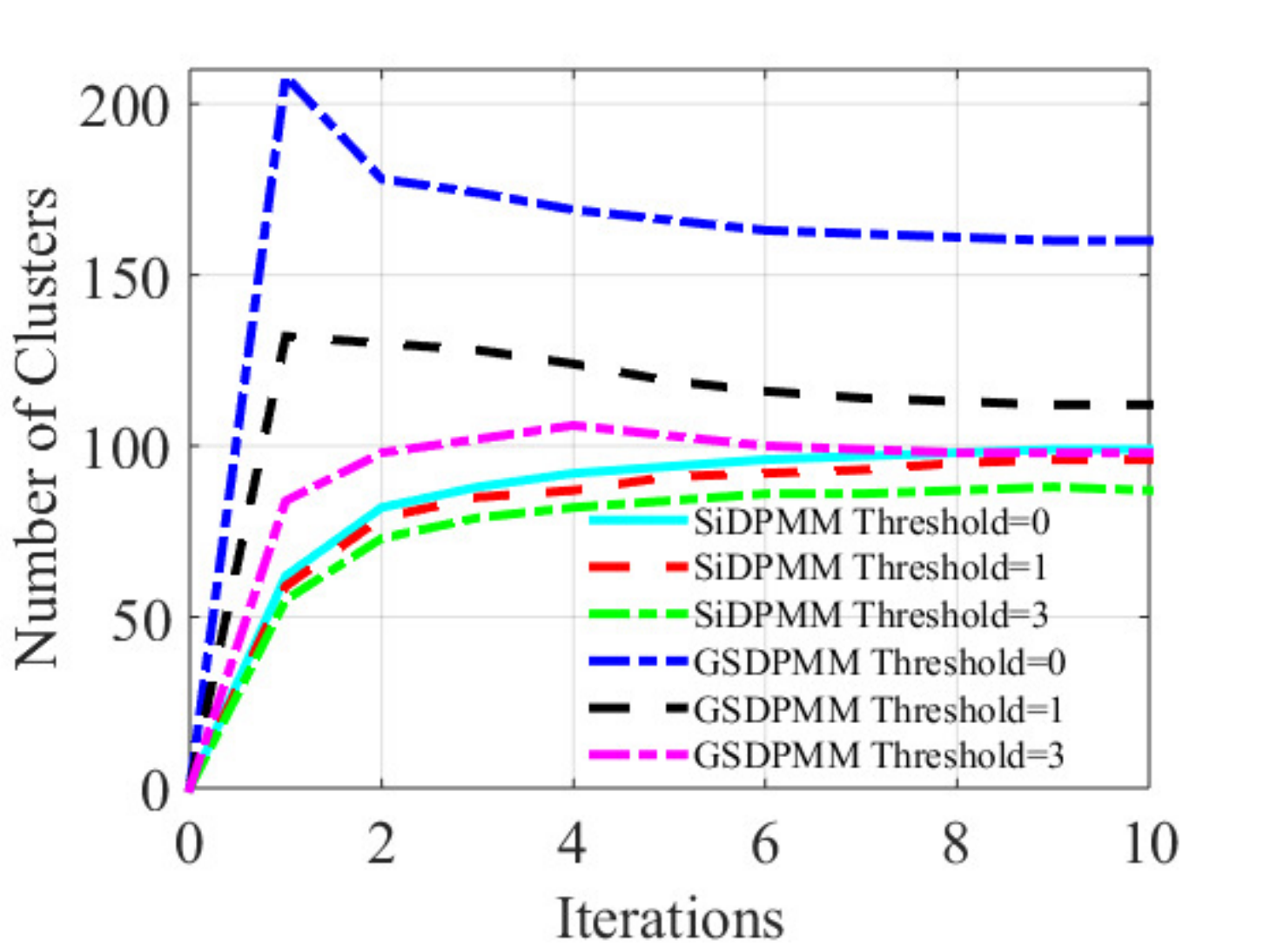}
		\caption{Tweet-Set}
		\label{Fig8_2}
	\end{subfigure}%
	
	\begin{subfigure}[b]{0.4\textwidth}
		\centering
		\includegraphics[width=\textwidth]{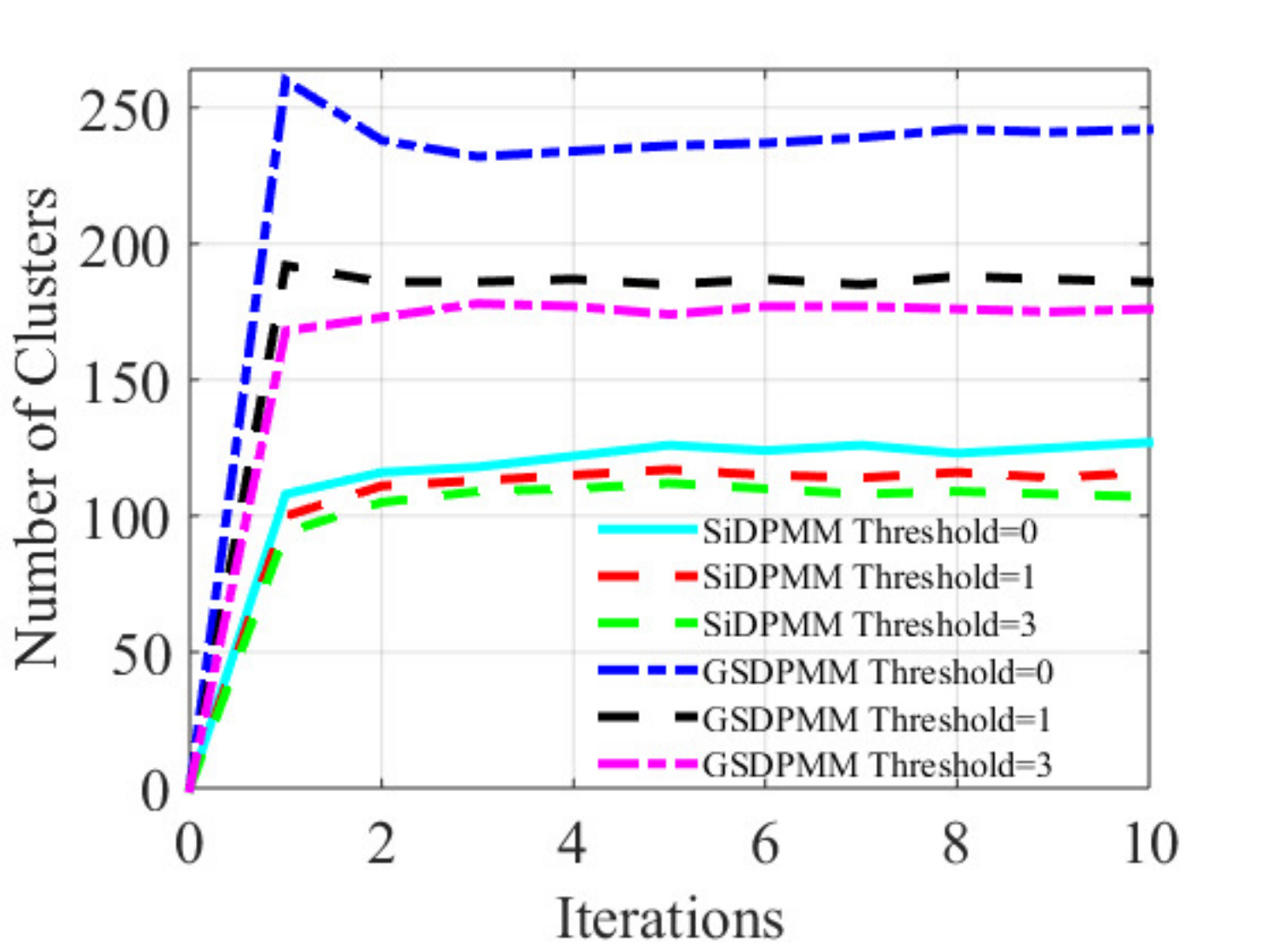}
		\caption{S-Set}
	\end{subfigure}%
	\begin{subfigure}[b]{0.4\textwidth}
		\centering
		\includegraphics[width=\textwidth]{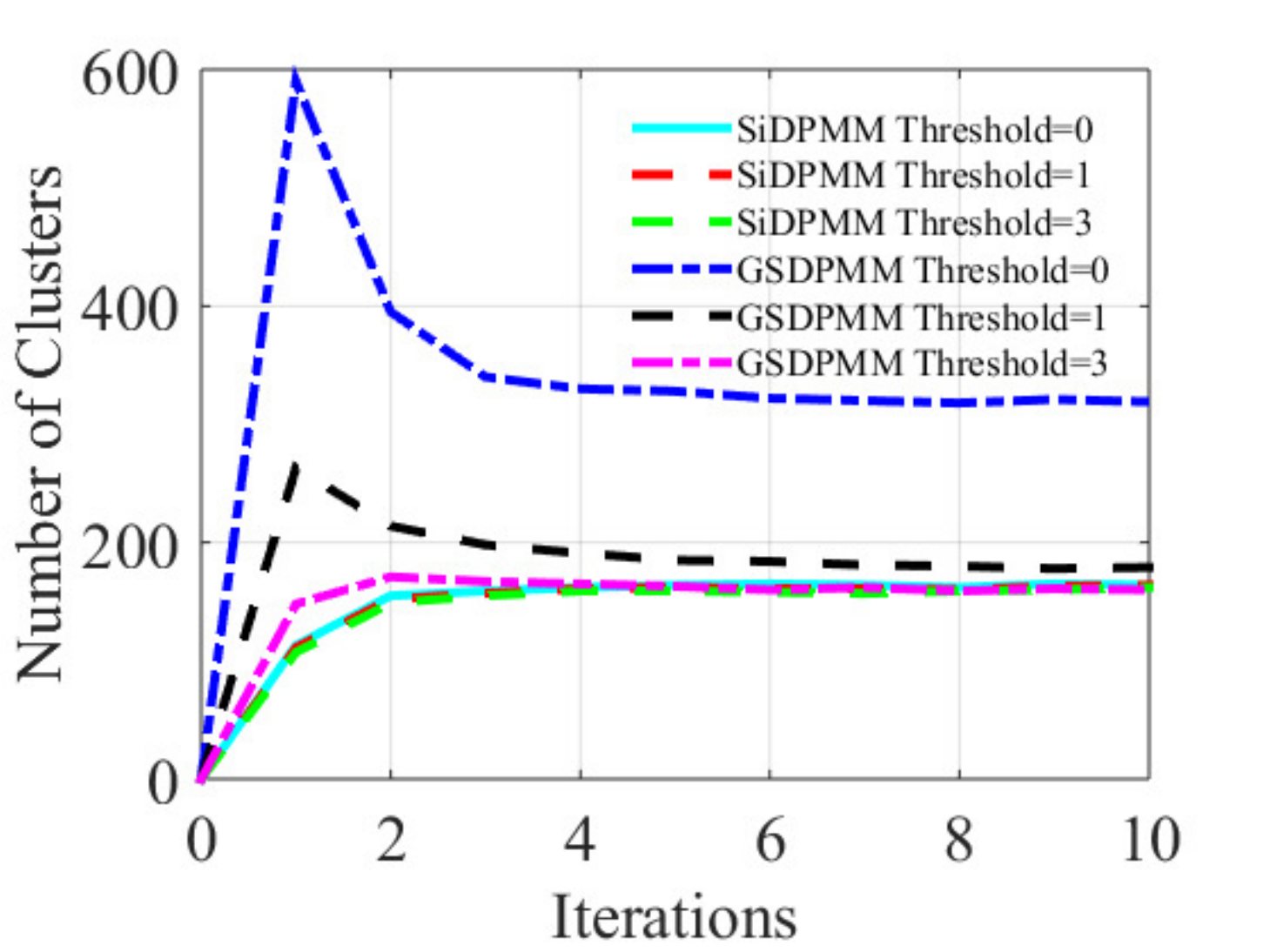}
		\caption{T-Set}
	\end{subfigure}%
	\caption{Number of clusters with size above a given threshold found in each iteration by SiDPMM and GSDPMM. 
		A cluster with size smaller than the given threshold does not count.
		Plots (a)-(d) are for the datasets 20NG, Tweet-Set, S-Set and T-Set respectively. 
	}
	\label{fig:small_clusters}
\end{figure}

\begin{figure}
	\begin{floatrow}
		\capbtabbox{%
			\scalebox{0.72}{
				\begin{tabular}{lrrr|rr}
					\toprule
					& \multicolumn{3}{c|}{Number of Clusters} & \multicolumn{2}{c}{Diff. Ratio} \\
					\midrule
					& \multicolumn{1}{c}{Ground Truth} & \multicolumn{1}{c}{GSDPMM} & \multicolumn{1}{c|}{SiDPMM} & \multicolumn{1}{c}{GSDPMM} & \multicolumn{1}{c}{SiDPMM} \\
					\midrule
					20NG  & 20    & 52    & 31    & 160\% & \textbf{55\%}\\
					T-Set & 152   & 323   & 171   & 113\% & \textbf{13\%} \\
					S-Set & 152   & 246   & 126   & 62\%  & \textbf{17\%} \\
					Tweet & 110   & 161   & 99    & 46\%  & \textbf{10\%} \\
					\bottomrule
				\end{tabular}%
			}
		}{%
			\caption{Inferred number of clusters by SiDPMM and GSDPMM. 
				Other baseline methods are not included because they require pre-specified number of clusters.}%
			\label{tab:num_of_clusters}
			
		}
		\ffigbox{%
			\includegraphics[width=0.35\textwidth]{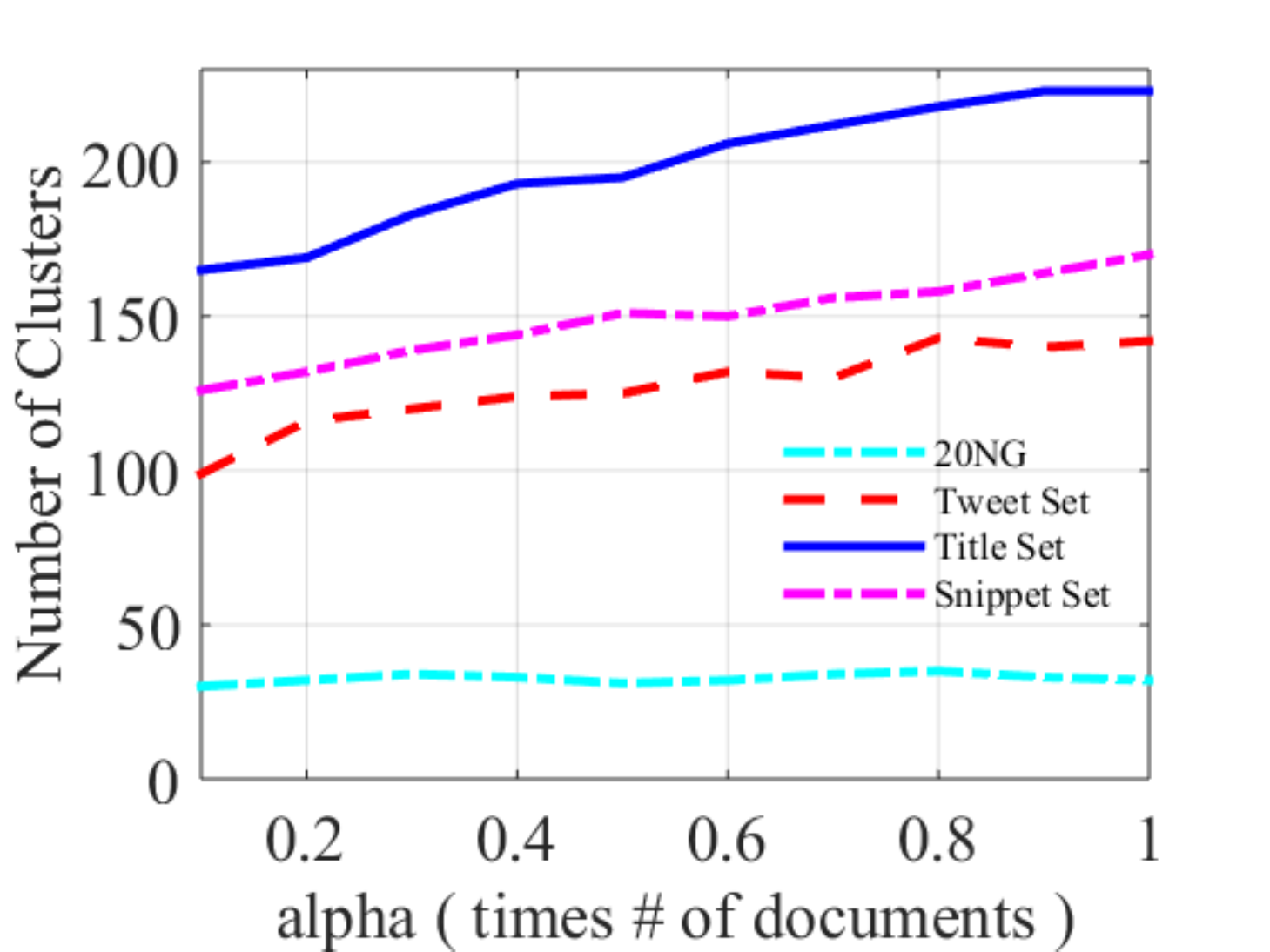}
		}{%
			\caption{Number of clusters found by SiDPMM with different $\alpha$ values, revealing the relative strength of prior (compared to likelihood) in determining posterior distribution}%
			\label{fig:alpha}
		}
	\end{floatrow}
\end{figure}

SiDPMM and GSDPMM can automatically determine the number of clusters.
Table~\ref{tab:num_of_clusters} shows that number of clusters inferred by 
SiDPMM are much more accurate compared to those from GSDPMM across all the datasets.
We can observe that GSDPMM tends to create more clusters than SiDPMM.
As illustrated in \figref{fig:small_clusters}, many of those clusters created by GSDPMM are quite small;
while in constrast,  SiDPMM tends to suppress tiny clusters and thus are more robust to outliers.
The sequential and word embedding components in SiDPMM are responsible for this regularization effect on number of clusters. 
 


The hyper-parameter $\alpha$ in the Dirichlet process determines the prior probability of creating a new cluster (see eq. \eqref{eq:11}). 
We explore the influence of different $\alpha$ values on our model. 
\figref{fig:alpha} shows that the number of clusters typically grows with $\alpha$;
as observed for Tweet Set, T-Set and S-Set,
but not the case for the 20NG dataset.
This reveals the relative strength of prior (compared to likelihood) in determining posterior cluster distribution. 
The documents in 20NG have large average length (137.5 words per document). 
In the sampling process, 
the likelihood dominates the posterior distribution 
and the small difference caused by different $\alpha$ in the prior distribution is negligible, 
while for documents with small average length, 
the difference in likelihood is not significant and thus prior affects more of the posterior distribution.

\section{Conclusion} 

In this paper, we propose a nonparametric Bayesian text clustering method (SiDPMM) 
which models documents as the joint of bag of words, word embeddings and sequential features. 
The approach is based on the observation that sequential information plays a key role in the interpretation of phrases 
and word embedding is very effective for measuring similarity between synonyms. 
The sequential features are extracted with an encoder-decoder component and word embeddings are extracted with the CBOW model. 
A detailed collapsed Gibbs sampling algorithm is derived for the posterior inference. 
Experimental results show our approach outperforms current state-of-the-art methods,
and is more accurate in inferring the number of clusters with the desirable regularization effect on tiny scattered clusters.

\bibliographystyle{splncs03}
\bibliography{DPMM}

\end{document}